\documentclass[conference]{IEEEtran}
\IEEEoverridecommandlockouts

\usepackage{times}

\usepackage[numbers]{natbib}
\usepackage{multicol}
\usepackage[bookmarks=true]{hyperref}
\usepackage{microtype}
\usepackage[table]{xcolor}
\usepackage{graphicx}
\usepackage{booktabs} 
\usepackage{tikz}
\usepackage{textcomp}
\usepackage{float}
\usepackage{algorithm}
\usepackage{algorithmic}
\usepackage{multicol}
\usepackage{multirow}
\usepackage{subfigure}
\usepackage{amsmath} 
\usepackage{enumerate}
\usepackage{amssymb}
\usepackage{amsbsy}
\usepackage{color}
\usepackage{hyperref}

\hypersetup{
    colorlinks=true,
    urlcolor=blue,
    filecolor = blue,
    citecolor = blue,
}

\begin{document}

\title{MQA: Answering the Question via Robotic Manipulation}
\author{Yuhong Deng$^\dagger$, Di Guo$^\dagger$, Xiaofeng Guo,  Naifu Zhang, Huaping Liu$^*$ and Fuchun Sun\\
Department of Computer Science and Technology, Tsinghua University, China
\thanks{$^\dagger$ indicates the authors with equal contributions. The authors are also with Beijing National Research Center for Information Science and Technology. Y. Deng is also with the Center of Intelligent Control and Telescience, Tsinghua Shenzhen International Graduate School, Tsinghua University.
This work was jointly supported by the National Natural Science Fund for Distinguished Young Scholars (62025304), and in part by the Seed Fund of
Tsinghua University (Department of Computer Science and Technology)-Siemens Ltd., China Joint Research Center for Industrial Intelligence and
Internet of Things.
$^*$Corresponding author: hpliu@tsinghua.edu.cn}%
}



%

\maketitle

\begin{abstract}
In this paper, we propose a novel task, Manipulation Question Answering (MQA), where the robot performs manipulation actions to change the environment in order to answer a given question. To solve this problem, a framework consisting of a QA module and a manipulation module is proposed. For the QA module, we adopt the method for the Visual Question Answering (VQA) task. For the manipulation module, a Deep Q Network (DQN) model is designed to generate manipulation actions for the robot to interact with the environment. We consider the situation where the robot continuously manipulating objects inside a bin until the answer to the question is found. Besides, a novel dataset that contains a variety of object models, scenarios and corresponding question-answer pairs is established in a simulation environment. Extensive experiments have been conducted to validate the effectiveness of the proposed framework. 


\end{abstract}

\IEEEpeerreviewmaketitle

\section{Introduction}
\label{section: intro}

\begin{figure}
	\centering
	\includegraphics[width=\linewidth]{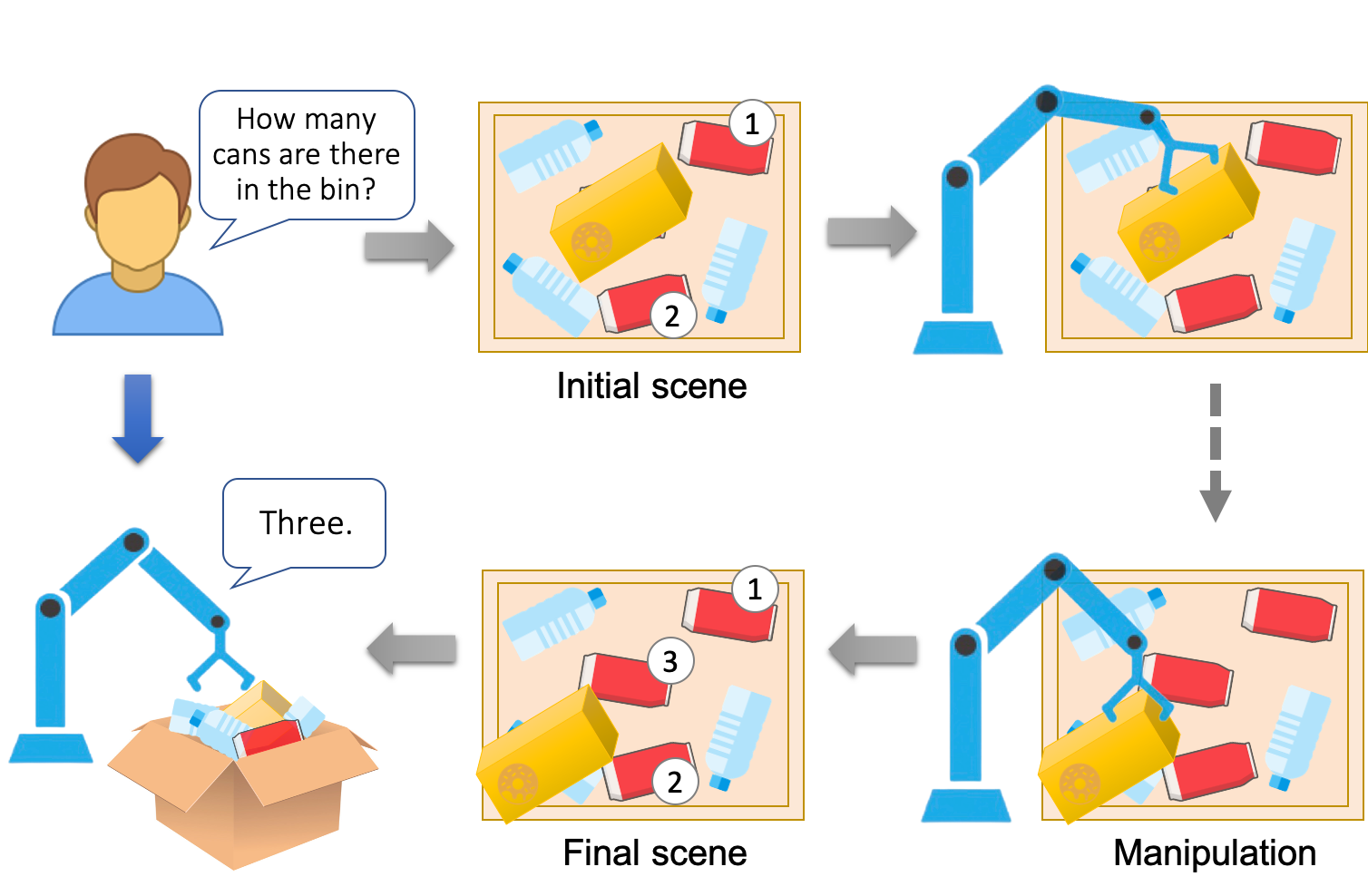}
	\caption{Given a question, the manipulator explores the bin with a series of manipulation actions to find the answer. In this example, two cans are directly visible in the initial scene, while there may also be some cans occluded by other objects. Therefore, the manipulator tries to push a food box to the side and another can is revealed. After exploring all possible areas, three cans are found at last. And the robot report the answer ``Three'' to the user.}
	\label{fig:summary}
\end{figure}

People have long anticipated the day when humans can ask questions to an intelligent robot directly with natural language and the robot knows to interact with the environment to respond. Imagine there is a bin in your kitchen which contains a variety of items, and you would like to know how many cans are left in it so that you can decide whether some replenishment should be done. Then you call your assistant robot in the kitchen and ask \textit{``How many cans are there in the bin?''} Having the question well understood, the robot starts to explore the bin, where all kinds of objects may be mixed together. As some cans may be occluded by some other objects and can not be seen directly, the robot has to generate a sequence of manipulation actions to change the current scenario in order to explore the bin thoroughly. As shown in Fig.\ref{fig:summary}, the robot keeps exploring the bin until all possible areas are explored and then it is able to report the answer to the user.

Recently, the task of Question Answering (QA) has attracted increasing attention from many researchers worldwide. In the big family of QA research, the popular QA chatbot tries to communicate with humans by scraping the Internet or database to get the answer to the question \cite{abdul2015survey}. One of the representative tasks among it is the Visual Question Answering (VQA) task \cite{anderson2018bottom}\cite{antol2015vqa}\cite{jang2017tgif}\cite{johnson2017clevr}, where the robot is required to have the ability to reason about the visual content in order to answer a question about the given visual input. As a step forward to realize the natural human-robot interaction, a much more challenging task, Visual Dialog, is proposed, where the robot needs to answer a coherent series of questions to the visual content \cite{das2017visual}\cite{visdial_diversity}. However, they only try to answer the question passively from the visual input and the robot's ability to move in the environment is ignored.

In the real-world environment, the perception should never be passive but an active process \cite{bajcsy1988active}\cite{bohg2017interactive}. Considering the embodiment of intelligent agents, next emerges a body of work on Embodied Question Answering (EQA), where the mobile robot is able to actively explore the environment to find the answer to the question \cite{das2017embodied}\cite{das2018neural}. In the EQA task, the robot needs to understand the acquired visual information and perform a series of actions accordingly to actively explore the environment to answer the question. A most important characteristic of the EQA task is that the perception and action ability of the agent are combined together. Additionally, under the large scope of EQA, Gordon et al. propose an Interactive Question Answering (IQA) task \cite{gordon2017iqa}, which points out that besides merely navigating the environment, the robot should also be able to execute some interactive actions based on the object's affordance, such as opening the door of the refrigerator to better find the answer to the question. But they are limited to some simple standard actions, and lack of manipulation.

In the real world environment, it is far more complex and highly unstructured. For example, in the cluttered scene, a target object may be occluded by other objects, which results in an even higher requirement on the robotic manipulation ability. To tackle this problem, we propose a novel task of Manipulation Question Answering (MQA), where the robot is required to find the answer by performing manipulation actions to actively explore the environment, rather than simply doing some predefined actions for the interaction. A comprehensive comparison of VQA, EQA, IQA, and the proposed MQA tasks is illustrated in TABLE \ref{cmp}. It can be seen that the VQA task only requires the agent to have the ability to understand the environment. Comparing to VQA task, the EQA task makes a big improvement by further leveraging the embodiment ability of the agent. The agent needs to explore the environment to find the answer. IQA task is an extension of the EQA task which also allows the interaction with the environment. And the proposed MQA contains all the characteristics of aforementioned tasks. Additionally, in the MQA task, the agent can perform manipulation actions to change the environment in order to answer the given question.

Meanwhile, the MQA task we proposed poses several new challenges. First, the robot is expected to perform manipulation actions to change the environment in order to find the answer, instead of merely referencing the static environment. And then, a new set of metrics is required to evaluate this new task as currently available research lacks quantitative accuracy metrics and benchmarks for the proposed task. Besides, there is no existing dataset suitable for our MQA task.

\begin{table}[t]
	\centering
	\caption{Comparison Among VQA, EQA, IQA and MQA}
	\begin{tabular}{c|c|c|c|c}
		\hline
		& VQA\cite{antol2015vqa}\cite{das2017visual} & EQA\cite{das2017embodied} & IQA\cite{gordon2017iqa} & MQA (Ours) \\ \hline
		Understanding  & $\checkmark$ & $\checkmark$ & $\checkmark$ &  $\checkmark$     \\ \hline
		Exploration & - & $\checkmark$ & $\checkmark$ & $\checkmark$ \\ \hline
		Interaction & - & - & $\checkmark$  & $\checkmark$        \\ \hline
		Manipulation & - & - & - & $\checkmark$         \\ \hline
	\end{tabular}
	\label{cmp}
\end{table}

In response to these challenges, we proposed a framework that integrates a QA module and a manipulation module to accomplish the newly defined MQA task. The contributions of this paper can be summarized as follows:
\begin{itemize}
	\item We formulate a novel Manipulation Question Answering (MQA) task and a solution framework is built to solve it.
	\item We design a Deep Q Network (DQN) for the robot to effectively generate manipulations for the MQA task.
	\item We build a novel MQA dataset including a variety of object models, bin scenarios and question-answer pairs. A corresponding benchmark is also established.
\end{itemize}

The organization of this paper is as follows. The related work is introduced in Section \ref{section:related}. We describe the MQA task in Section \ref{section: Problem}. Section \ref{section: dataset} includes the establishment of the MQA dataset and its analysis. The proposed MQA framework is presented in Section \ref{section: model}. Experimental results and analysis are demonstrated in Section \ref{section: experiment}. Finally, we come to the conclusion of the paper.

\section{Related Work}
\label{section:related}
\subsection{Robotic QA Application}
With the improvement of the robot's ability to perceive the environment, it has become a new trend to use robots to assist QA tasks. For example, the visual system of the robot is able to obtain much semantic information of the captured visual information, which provides additional clues for the questioning answering task. Additionally, the robot's motion ability brings the flexibility to move around, which makes the robotic QA task more meaningful in practical scenarios. Tan et al. use several agents to move around jointly in interactive environments to answer a question \cite{multiagent}. In \cite{he2017educational}, an educational robot system of VQA is proposed. The robot can detect objects in the scene and then generate relevant questions to ask the children. Considering the practical implementation of the VQA model, Sejnova et al. \cite{sejnova2019exploring} investigate the optimal viewpoint selection problem to maximize the performance of VQA model in real world. In \cite{nazarczuk2020shop}, a robotic visual reasoning benchmark which concludes question-answer pairs and other semantic information is also established to facilitate robotic applications. Actually, the type of robotic QA task is not only a simple natural language processing task. It is supposed to combine the  QA task, the action and perception ability of robot together, which is greatly more challenging than the normal QA task.

\subsection{Robotic Manipulation in Clutter}
The task of robotic manipulation in a cluttered environment has been investigated for decades \cite{6672028}\cite{deng2019deep}\cite{7759839}\cite{doi:10.1177/0278364919868017}, while most work tries to find proper grasping locations to grasp an object from clutter without any cognitive purposes. In \cite{doi:10.1177/0278364919868017}, an affordance map is employed to generate a pixel-wise grasping point by analyzing the whole cluttered scene. Additionally, action primitives such as pushing and picking are cooperated to solve the problem of grasping an occluded object from a clutter \cite{zeng2018learning}. In \cite{danielczuk2019mechanical}, a mechanical search policy is proposed to retrieve occluded target objects from cluttered bin using parallel grasping, suction grasping and pushing action primitives. The objective of current work mostly focuses on how to generate a good manipulation to grasp target objects from clutter. Recently, complex robotic tasks in unknown or unstructured environments tend to be the combining of perception, control, and cognition \cite{combined1}\cite{combined2}. The objective of the proposed MQA task further involves a cognitive purpose, which require the robot to generate a sequence of manipulation actions to explore the environment and answer people’s questions.

\subsection{Semantic Understanding in Robotic Manipulation}
The semantic understanding of the environment plays a rather important role in the task of robotic manipulation, which is investigated by a lot of research work recently. Kenfack et al. propose a RobotVQA system, which can generate a semantic scene graph of the observed scenario. The robot can then resort to the scene graph to manipulate the object more efficiently \cite{kenfack2020robotvqa}. In a canonical task, rearrangement, the robot is required to change a given environment to a specified state by manipulation given different sources of information such as object poses, images, language description to understand the environment \cite{batra2020rearrangement}. Tan et al. propose the new Embodied Scene Description problem, where the agent needs to find an optimal viewpoint in its environment for scene description tasks \cite{viewpoint}. Additionally, Gandhi et al introduce the sound information for a richer environment understanding. The interaction between the sound and robotic action is investigated \cite{gandhi2020swoosh}. In the proposed MQA task, the robot is also required to have a semantic understanding of the environment. Meanwhile, the robot should well understand the question and thus perform manipulation actions to respond.

\section{PROBLEM FORMULATION}
\label{section: Problem}

In this section, we formally state the problem formulation of the MQA task. Given a question $q$, which is represented by a natural language sentence, and the state vector $s_t$ of the environment, where $t$ is the time instant, the task of MQA is to obtain a manipulation action
\[
a_t = \mathcal{M}(s_t, q),
\]
where $\mathcal{M}$ is the manipulation module that should be developed. This forces the state to become as
\[
s_{t+1} = f(s_t, a_t).
\]
where $f$ is the intrinsic dynamics of the environment.

The above procedure should be repeated until the $stop$ action is triggered. Then the answer corresponding to the problem $q$ can be obtained as
\[
ans = {\mathcal{QA}}(s_0,s_T, q),
\]
where $T$ is the terminated time instant, $s_0$ is the initial state and $s_T$ is the terminal state. The QA model $\cal{QA}$  adopts the two states and the inquiry $q$ to generate the answer.

In this work, our task is to design the manipulation module $\mathcal{M}$ and the QA model $\mathcal{QA}$, and integrate them into a unified framework to realize the MQA. The challenge of the MQA task mainly lies in two aspects. One is that the robot is required to understand the semantic information of the given question and visual environment. The other is that a manipulation action sequence should be generated based on this high-level semantics to answer the question.

\section{MQA DATASET}
\label{section: dataset}
\begin{figure}[]
	\centering
	\subfigure[simulation environment]{
		\begin{minipage}[]{0.45\linewidth}
			\centering
			\includegraphics[height = 0.8 in]{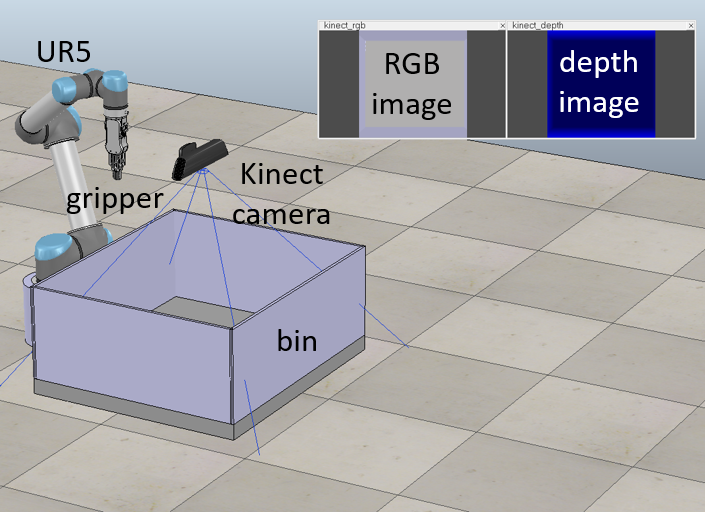}
		\end{minipage}%
	}%
	\subfigure[3D object models]{
		\begin{minipage}[]{0.45\linewidth}
			\centering
			\includegraphics[height = 0.8 in]{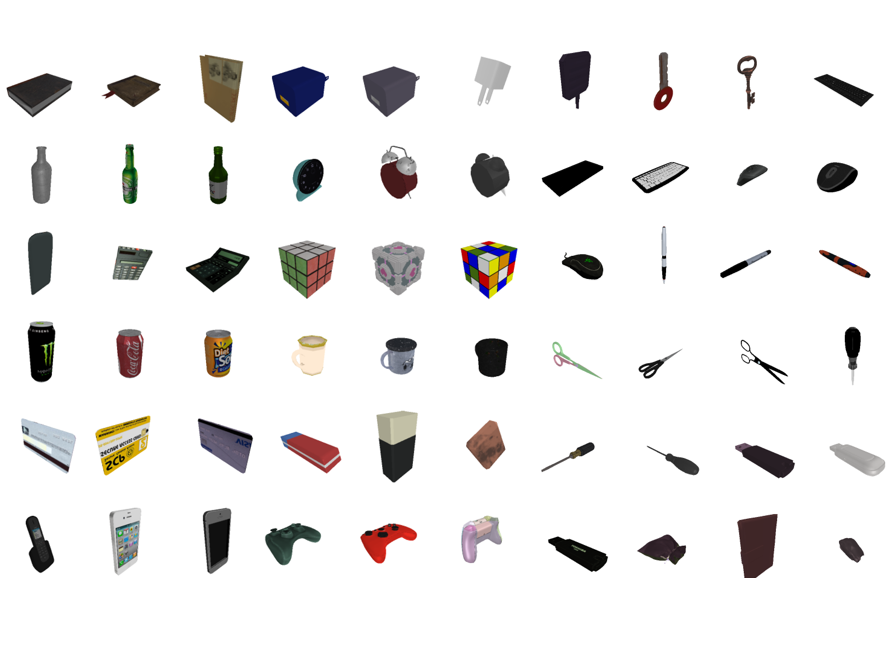}
			\label{fig:objs}
		\end{minipage}%
	}%
	\\
	\subfigure[different scenes]{
		\begin{minipage}[]{0.45\linewidth}
			\centering
			\includegraphics[height =0.8 in]{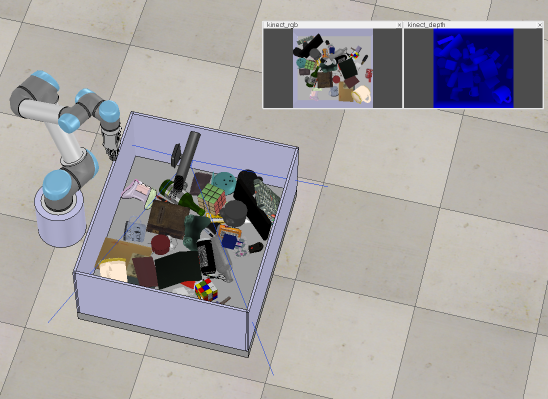}
	\end{minipage} }%
	\subfigure[question-answer pairs]{
		\begin{minipage}[]{0.45\linewidth}
			\centering
			\includegraphics[height=0.8 in]{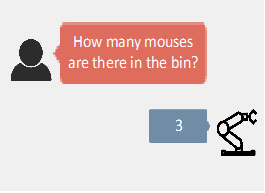}
		\end{minipage}%
	}%
	\centering
	\caption{The overview of the MQA dataset.}
	\label{fig:dataset}
\end{figure}

As our task is newly proposed, there is no suitable off-the-shelf dataset for experiments. Therefore, we establish our own dataset for the MQA task. The MQA dataset is built under the V-REP \cite{rohmer2013v} simulation environment, and it is composed of a variety of 3D object models, different scenes, and corresponding question-answer pairs for every scene (Fig.\ref{fig:dataset}). The established MQA dataset is designed with the explicit goal of training and testing how a robotic manipulator actively explores the environment with manipulation actions to find the answer to the given question. The dataset is already released and publicly available.

\subsection{Simulation Environment}
In the MQA task, the robotic manipulator needs to perform manipulation actions to change the environment in order to answer the question. Considering the situation that the real robotic platform is usually subjected to restricted experimental environment, it is thus not scalable to the unstructured real environment. Additionally, it is very costly and unsafe to train learning algorithms that require thousands of iterations with the real robot. Therefore, we resort to the V-REP robotic simulator, which provides satisfying realistic visual renderings and an accurate physics engine (bullet engine) to generate the large-scale MQA dataset.

We consider the bin scenario, where selected objects are placed in the bin randomly. A UR5 robotic manipulator with a gripper is placed in front of the bin to perform manipulation actions. The visual information of the scene is captured with a Kinect camera, which is fixed on the top of the bin.

\subsection{3D Objects}
\begin{figure}[b]
	\centering
	\includegraphics[width=0.8\linewidth]{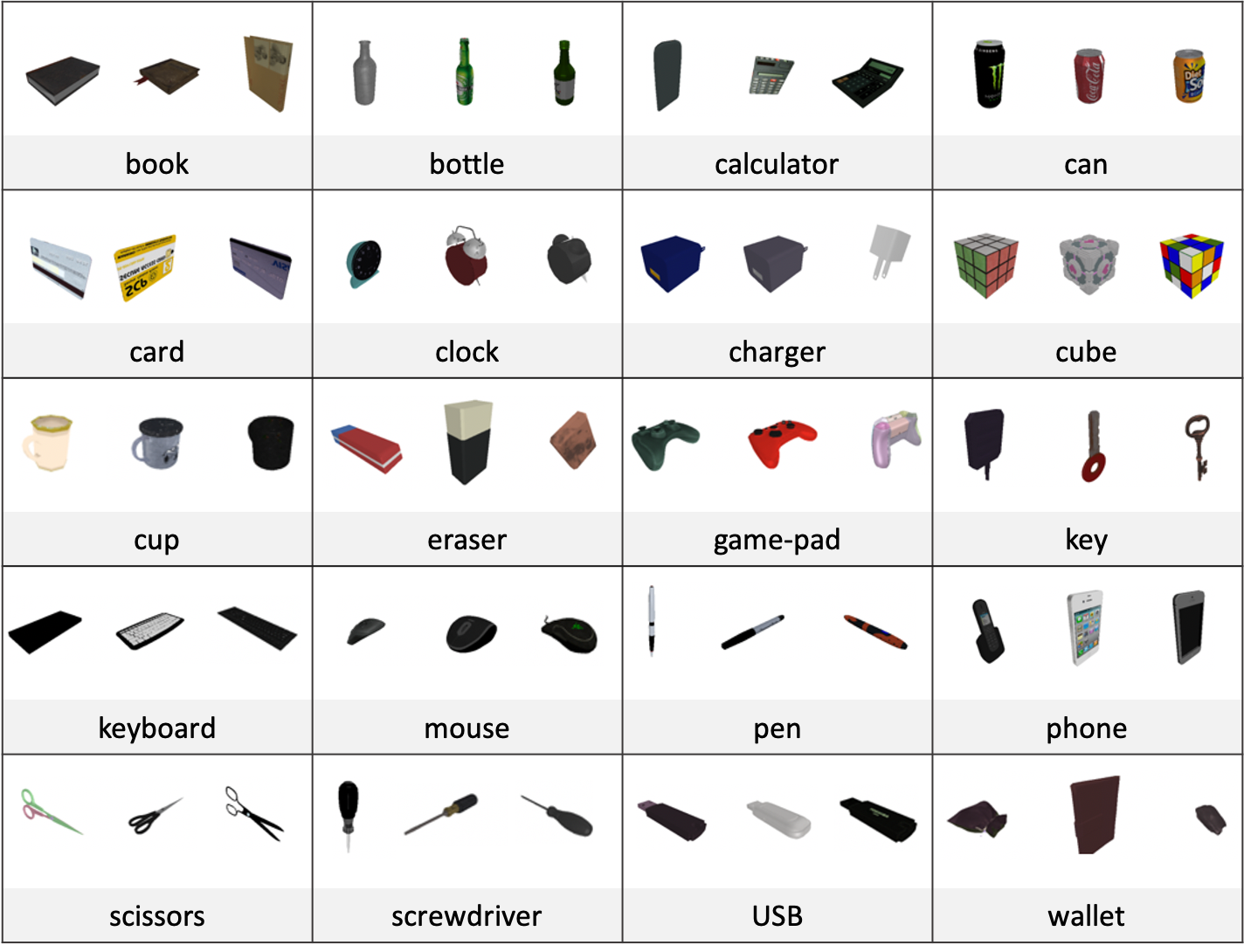}
	\caption{Objects selected in the MQA dataset.}
	\label{fig:obj_tab}
\end{figure}
We select 20 classes of common objects (Fig.\ref{fig:obj_tab}), and 3 different instances are picked for each class. We try to make the size and color of these objects as diverse as possible. Some objects have relatively regular shapes, such as the box and cylinder, but some are very irregular, such as a gamepad, bringing challenges for the perception and manipulation. There are altogether 60 different instances in the dataset, and each instance is represented by its 3D-model and corresponding texture, which makes it look more real.



\subsection{Scene Generation}
\begin{figure}[h]
	\centering
	\subfigure[\textit{easy} scene]{
		\begin{minipage}[t]{0.32\linewidth}
			\centering
			\includegraphics[width=\linewidth]{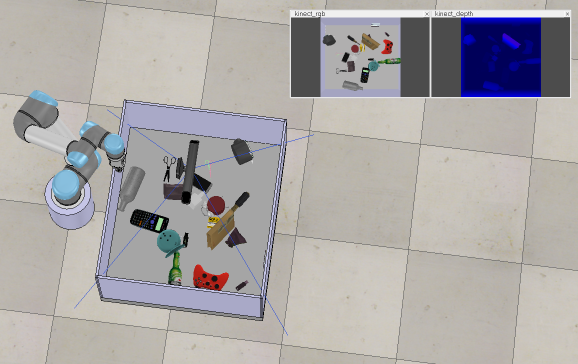}
		\end{minipage}%
	}%
	\subfigure[\textit{medium} scene]{
		\begin{minipage}[t]{0.32\linewidth}
			\centering
			\includegraphics[width=\linewidth]{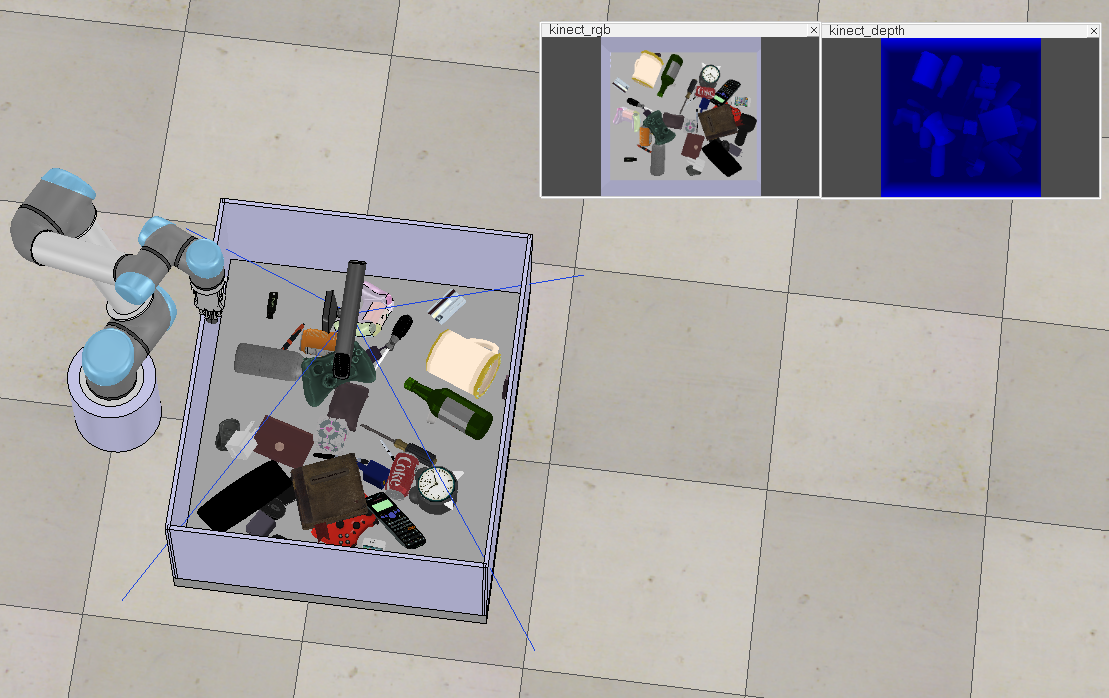}
		\end{minipage}%
	}%
	\subfigure[\textit{hard} scene]{
		\begin{minipage}[t]{0.32\linewidth}
			\centering
			\includegraphics[width=\linewidth]{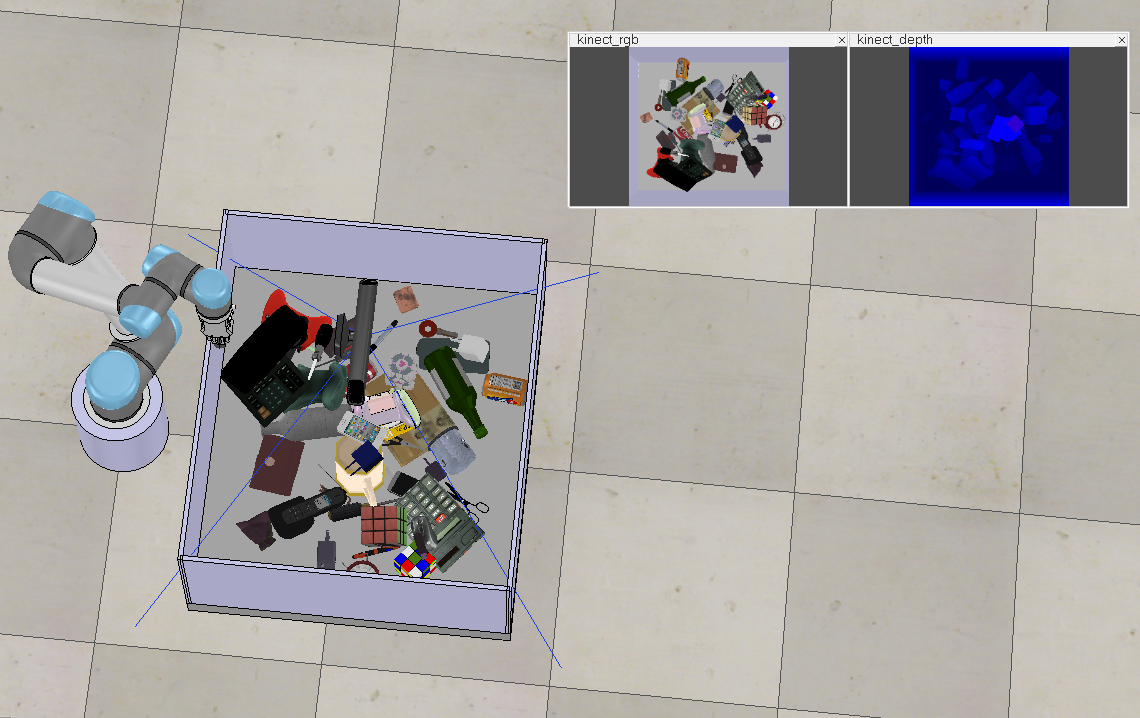}
		\end{minipage}
	}%
	\centering
	\caption{Initial scenes with different difficulty level}
	\label{fig:scene}
\end{figure}

The scenes are generated according to different difficulty levels. For each scene, We randomly select $n$ objects from all 60 objects, where $n$ is determined by the difficulty level of the scene. We define three difficulty levels, namely \textit{easy} ($n=20$), \textit{medium} ($n=35$) and \textit{hard} ($n=50$) levels. To generate the scene, the objects are thrown into the bin randomly and the position of the object is recorded. Fig.\ref{fig:scene} demonstrates some initial scenes regarding to different difficulty levels. We generate 100 initial scenes, among which 30 are of \textit{easy} level, 30 are of \textit{medium} level, and 40 are of \textit{hard} level. It's noted that these 100 initial scenes will be changed continuously during the manipulation process resulting in 100 scene series.




\subsection{Question-Answer Pairs Generation}
\label{sec:question}
The types of questions are designed considering the bin scenario. Inspired by \cite{das2017embodied}\cite{gordon2017iqa}, the template-based method is used to generated questions. And four kinds of questions are considered in the dataset, namely COUNTING, EXISTENCE, SPATIAL, and LOGIC questions. The question templates are demonstrated in TABLE \ref{table:questions}. The \textit{\{OBJ\}} can be replaced with specific object classes in Fig.\ref{fig:obj_tab}.

\begin{table}[H]
	\caption{Question Templates}
	\begin{tabular}{l|l}
		\hline
		COUNTING        &  \textit{How many \{OBJ\} are there in the bin?}\\\hline
		EXISTENCE      &  \textit{Is there a \{OBJ\} in the bin?}\\\hline
		SPATIAL  &  \textit{Is there a \{OBJ1\} under the \{OBJ2\}?}\\\hline
		LOGIC & \textit{Is there a \{OBJ1\} and a \{OBJ2\} in the bin?}\\\hline
		
	\end{tabular}
	\label{table:questions}
\end{table}

These generated questions cover a wide variety of situations, requiring different degrees of interaction between the robot and environment. For example, COUNTING questions would necessitate the robot having a memory, in order not to double count a query object. Besides, multiple manipulations may be required to answer EXISTENCE questions, as the robot has to actively explore the entire bin when the query object is not directly visible at first. SPATIAL questions require the robot to identify the spatial relationship between two objects.

Based on the templates, we generate very diverse and large amount of questions in the dataset. For each initial scene, 20 EXISTENCE questions, 20 COUNTING questions, 380 LOGIC questions and 380 SPATIAL questions are generated respectively. For training purpose, we randomly select 20 questions for each type. The answer to each question can be automatically generated from an oracle view in the simulator .

It is noted that since COUNTING question represents a typical type of question that usually requires the agent to perform multiple actions in order to answer, reflecting the core idea and characteristics of the proposed MQA task. It is specifically selected to be solved in this paper and we leave the others for the future work.

\subsection{Dataset Analysis}

The constructed MQA dataset is composed of 100 \textit{scene series}, each of which consists of one initial scene, and multiple successive scenes. There are 20 question-answer pairs for each type of questions. Therefore, 2000 questions are generated for each question type. In the experiment, the training set accounts for 70\% of the total data, and the test set accounts for 30\% of the total data.

\begin{figure}[t]
    \centering
    \subfigure[answers to COUNTING questions]{
    \begin{minipage}[t]{0.45\linewidth}
	\centering
	\includegraphics[width=1\linewidth]{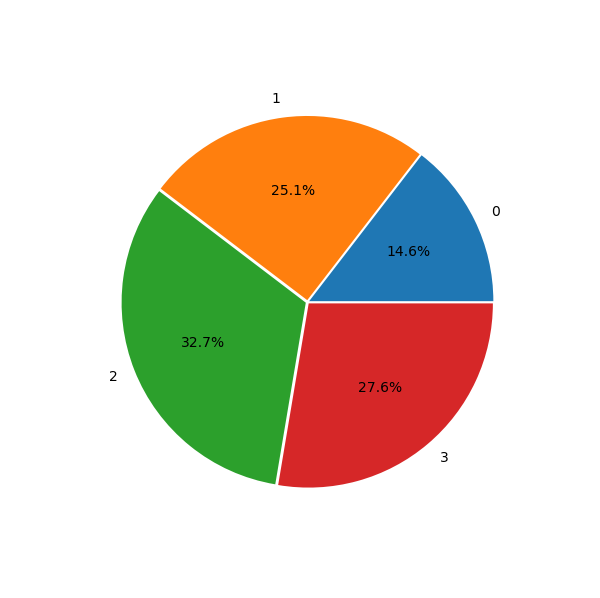}
	\label{fig:ans_count}
	\end{minipage}
    }
    \hspace{0.05in}
    \subfigure[answers to COUNTING questions in different scenes]{
    \begin{minipage}[t]{0.45\linewidth}
	\centering
	\includegraphics[width=1\linewidth]{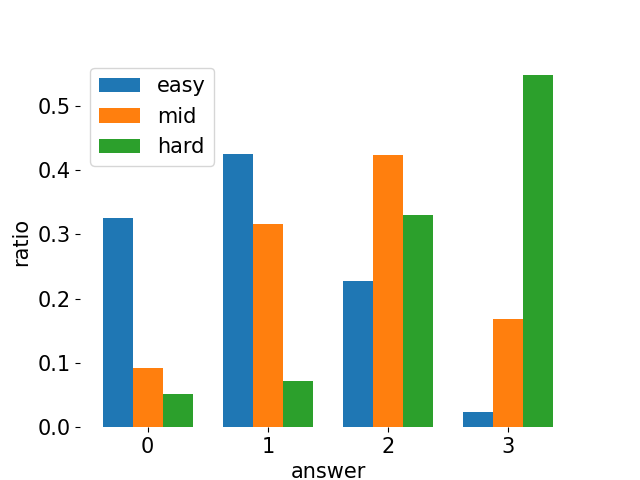}
	\label{fig:ans_count_diff}
	\end{minipage}
	}
	\centering
	\caption{COUNTING questions dataset analysis}
\end{figure}

Among all COUNTING questions, the answers range from 0 to 3. We find the probability of 0, 1, 2 and 3 answers are close to each other, and the dataset of COUNTING questions is well balanced (Fig.\ref{fig:ans_count}). At the same time, for scenes of different difficulty levels, the answers presents different distribution (Fig.\ref{fig:ans_count_diff}). In \textit{easy} scenes, most of the answers are 0 and 1. In \textit{medium} scenes, most of the answers are 1 and 2. In \textit{hard} scenes, most of the answers are 2 and 3. This is consistent with practical situation. As the more complex the scene is, the more objects there are. And it is more likely that objects are occluded by each other, bringing challenges to answer COUNTING questions.

For other types of questions, among EXISTENCE questions, we set the possibility of answer \textit{``Yes''} to be greater than that of \textit{``No''} in our dataset. This is because we consider that in reality, when asking EXISTENCE question, we often believe that there are books in the bin. This unbalanced distribution of answers makes it difficult to solve the EXISTENCE questions. Among SPATIAL questions, although we have a cluttered scene, due to the large number of object pairs in total, the \textit{``Yes''} answers only occupy a small ratio. Among LOGIC questions, the situation is similar to EXISTENCE questions, while we have more \textit{``No''} answer than it is in the EXISTENCE questions. It is because that the requirements for LOGIC questions are more strict.

\section{MQA MODEL}
\label{section: model}
\begin{figure}[h]
	\centering
	\includegraphics[width=\linewidth]{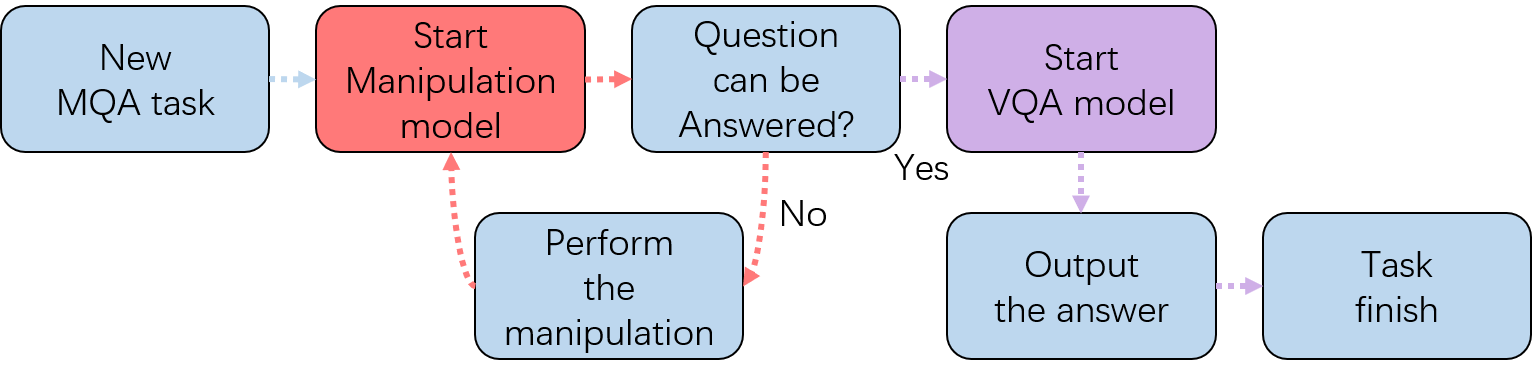}
	\caption{The workflow of MQA system.}
	\label{fig: system}
\end{figure}

The proposed MQA system is mainly composed of two parts, manipulation module and QA module. An overview of the workflow is demonstrated in Fig.\ref{fig: system}. When a new MQA task starts, the manipulation module will be activated first. The manipulation module will take the RGBD images of the scene and the question as input and output manipulation actions. The agent explores the environment until the question can be answered. The manipulation module decides when to stop exploring. Then, the QA module will give an answer based on the initial scene, the final scene and the question.

\begin{figure*}[h]
	\centering
	\includegraphics[width=0.95\linewidth]{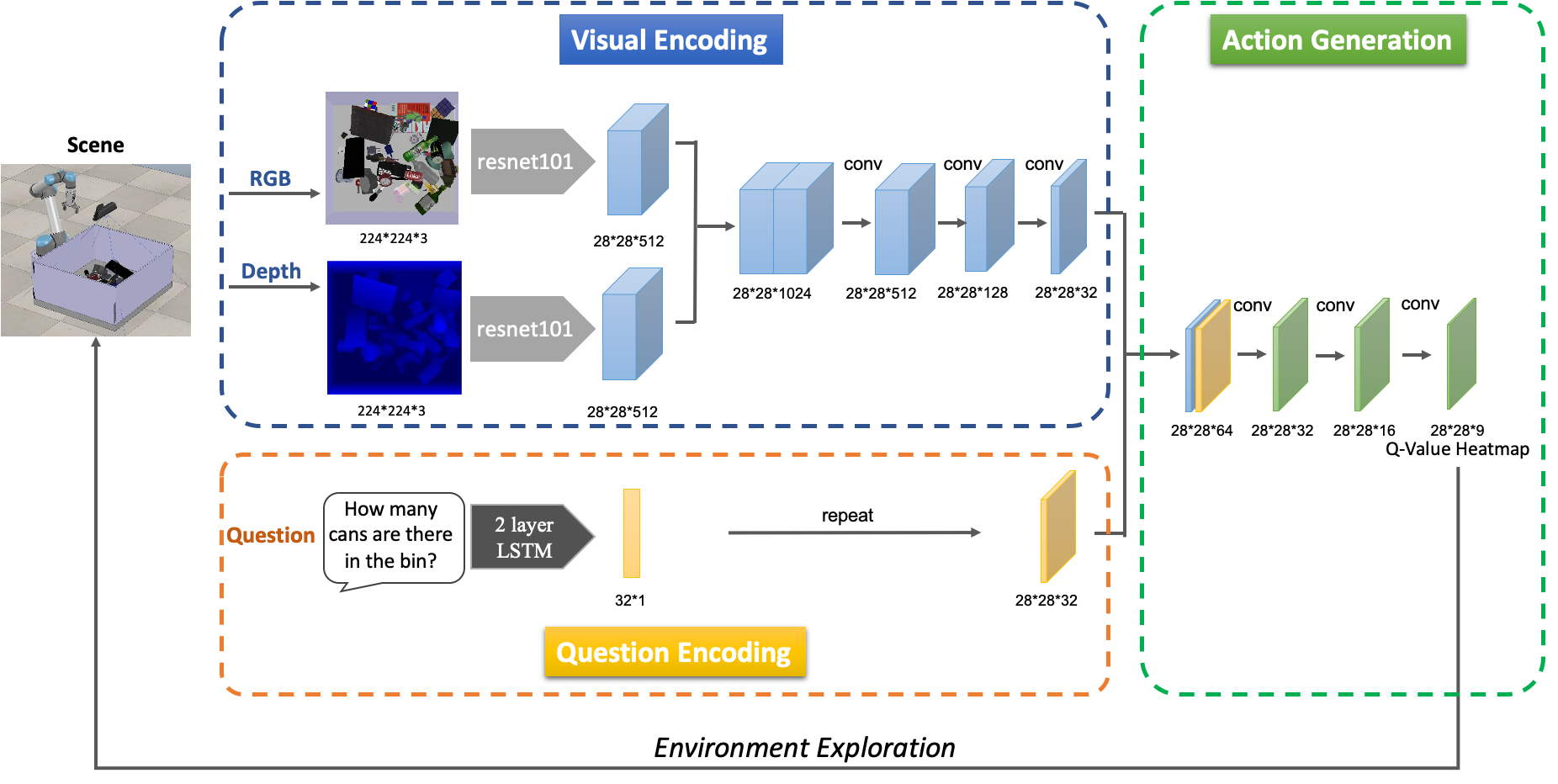}
	\caption{The architecture of the proposed manipulation module}
	\label{fig:action}
\end{figure*}

\subsection{Manipulation Module}
As demonstrated in Fig.\ref{fig:action}, the proposed manipulation module is composed of three parts, namely the visual encoding part, the question encoding part and the action generation part. A DQN model is designed to generated manipulation actions. We will elaborate the state space, action space, and reward design as follows.

\subsubsection{state space}
Our state space is a continuous space that comes from two sources: RGBD images and question. The outputs of the visual encoding part and question encoding part are concatenated providing the current state. The state in our system has a size of $28 \times 28 \times 64$. In the visual encoding part, both the RGB and depth images of the scene are fed into the network in parallel. We use the Resnet101 architecture pre-trained on the ImageNet dataset as the backbone of this part. The FC layer of the Resnet101 is replaced with a convolutional layer to output the feature map. The obtained feature map has a size of $28 \times 28 \times 32$, where each point represents an area in the RGB image. For the question encoding part, we use a 2-layer LSTM network to encode the given question. It is because LSTM is suitable for encoding natural language and it can be generalized to handle different types of questions.

\subsubsection{action space}
We use pushing actions for the robot to manipulate objects in the scene. We define 9 specific pushing actions on each location, the robot can choose to push the object from 8 directions with a fixed distance or to stop manipulation. We use $O_i = i*45^{\circ} (i=0, 1, 2, ..., 7)$ to represent 8 directions, and the push distance is 1/4 size of the image width.

\subsubsection{reward design}
A Q-value heatmap obtained from action generation part is used to design the reward for network updating. In order to make the model learn manipulation actions more efficiently, we design the reward of DQN model according to different conditions.

Firstly, we define the overlap rate $\chi$ of an object.
\begin{equation}
	\label{equation_2}
	\chi = 1-\frac{A_s}{A_t}
\end{equation}
where $O$ is the overlap rate, $A_t$ is the complete projected area of the object on the RGB image plane and $A_s$ is the projected area of the same object part that can be seen.

We design the reward based on the complexity of the scene. The more simple the scene is, the more easy it is to answer the question. In this paper, we take the COUNTING question as an example. We consider that the COUNTING question is most likely to be answered correctly when both all queried objects in questions can be seen in the scene and with an overlap rate of at most 0.2, which is supposed to be a simple scene. If the scene doesn't meet this requirement, the model will get a reward of -1 for doing harmful to the question answering task. Otherwise, the model will get a reward of 1.

If the manipulation model still outputs the push manipulation when the scene is simple enough, the model will receive a reward of 0 for outputting redundant manipulation. If the model outputs the push manipulation when the scene is not simple enough for question answering, the model will get a reward composed of two parts: $R_e$ denoting the reward for exploring the environment and $R_q$ denoting the reward for answering the questions. The total reward is equal to the weighted sum of the two parts (Eq.(\ref{equation_1})). The weight changes with time. At the beginning, the weight $\beta$ of the $R_e$ is 1. Because it is difficult to directly learn the manipulation that is beneficial to question answering in the initial stage, it is necessary to let the model learn to explore the environment first. With the increase of training times of DQN, $\beta$ decreases, and the model is supposed to learn exploration manipulations which are beneficial for question answering.
\begin{equation}
	\label{equation_1}
	Reward = \beta R_e +(1-\beta)R_q \quad 
\end{equation}
where $\beta$ is defined as
\begin{equation}
\label{equation_1_1}
\beta=
\begin{cases}
1-0.5\frac{S_n}{T} & S_n \leq T\\
0.5 & S_n > T
\end{cases}
\end{equation}
where $S_n$ is the step numbers of DQN during training, $T$ is the half size of replay memory.

$R_e$ is defined according to the following steps: $\boldsymbol{p_1},\boldsymbol{p_2},...\boldsymbol{p_n}$ denote positions of objects in the scene. Positions will be updated to $\boldsymbol{p_1^{'}},\boldsymbol{p_2^{'}},...\boldsymbol{p_n^{'}}$ after a manipulation action is implemented. Then $R_e$ is defined as:
\begin{equation}
	\label{equation_2}
	R_e = max(\frac{\vert\boldsymbol{p_i}-\boldsymbol{p_i^{'}}\vert} {\vert \boldsymbol{p_i} \vert},i = 1,2,3....n)
\end{equation}

For $R_q$, since there may be more than one target object in the scene, it is designed to have two parts: global part $R_g$ and local part $R_l$. We consider objects of the target category which is the object category mentioned in the question. Suppose there are $m$ objects of target category in the scene ($m =0$ means that the scene is simple enough and no target object is in it), $(\chi_1,\chi_2,...,\chi_m)$ is the overlap rate of these $m$ objects of target category, $(\chi_1^{'},\chi_2^{'},...,\chi_m^{'})$ is the overlap rate of them after a manipulation action is implemented. $R_g$ is the rate of change of average overlap rate of all objects:
\begin{equation}
	\label{equation_3}
	R_g = \frac{\chi_a-\chi_a^{'}}{\chi_a}, where \ \chi_a = \frac{\sum_{i=1}^{m}\chi_i}{m}, \chi_a^{'} = \frac{\sum_{i=1}^{m}\chi_i^{'}}{m}
\end{equation}

As for $R_l$, we pay attention to a single object. And we focus on the most occluded object. Because the most occluded object is most difficult to be detected by the algorithm, which is likely to lead to wrong results.
\begin{equation}
\begin{aligned}
	\label{equation_7}
	R_l = \frac{max(\chi_1,\chi_2,...,\chi_m)-max(\chi_1^{'},\chi_2^{'},...,\chi_m^{'})}{max(\chi_1,\chi_2,...,\chi_m)} \\	
\end{aligned}
\end{equation}

Therefore, $R_q$ can be represented by the weighted sum of $R_g$ and $R_l$, and we will study the influence of $R_g$ and $R_l$ respectively in Section \ref{ablation_study}.
\subsection{Question Answering Model}
\begin{figure}[h]
	\centering
	\includegraphics[width=\linewidth]{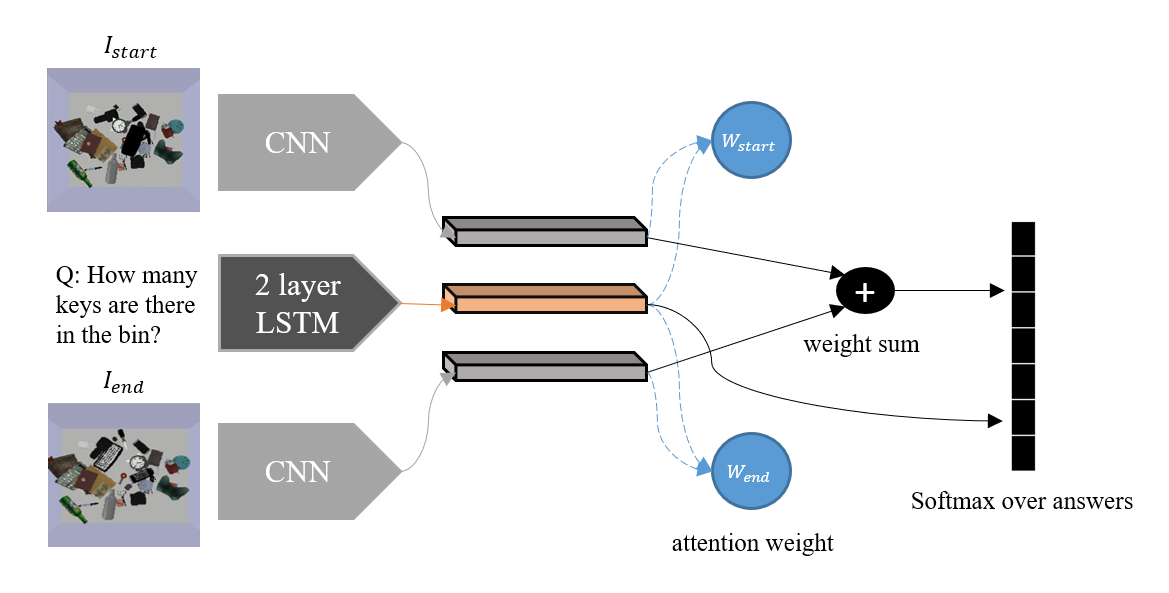}
	\caption{VQA model will give an answer based on the initial RGB image, the final RGB image and the question. }
	\label{fig:vqa}
\end{figure}

We use the VQA model in \cite{das2017embodied} for the question answering model. The question answering model will be executed when a $stop$ action is generated by the manipulation module or a maximum number of manipulation steps is achieved. We denote the RGB image before manipulations as $I_{start}$ and the RGB image after manipulations as $I_{stop}$. As shown in Fig.\ref{fig:vqa}, these two images are encoded by convolutionary networks, and the question is encoded by a 2-layer LSTM network. The image-question similarity between the question and the two images will be calculated as attention weights to fuse the two image features. Then, the attention-weighted image features combined with question encoding are passed through a softmax classifier to predict the answer.

\begin{figure*}[h]
	\centering
	\includegraphics[width=\linewidth]{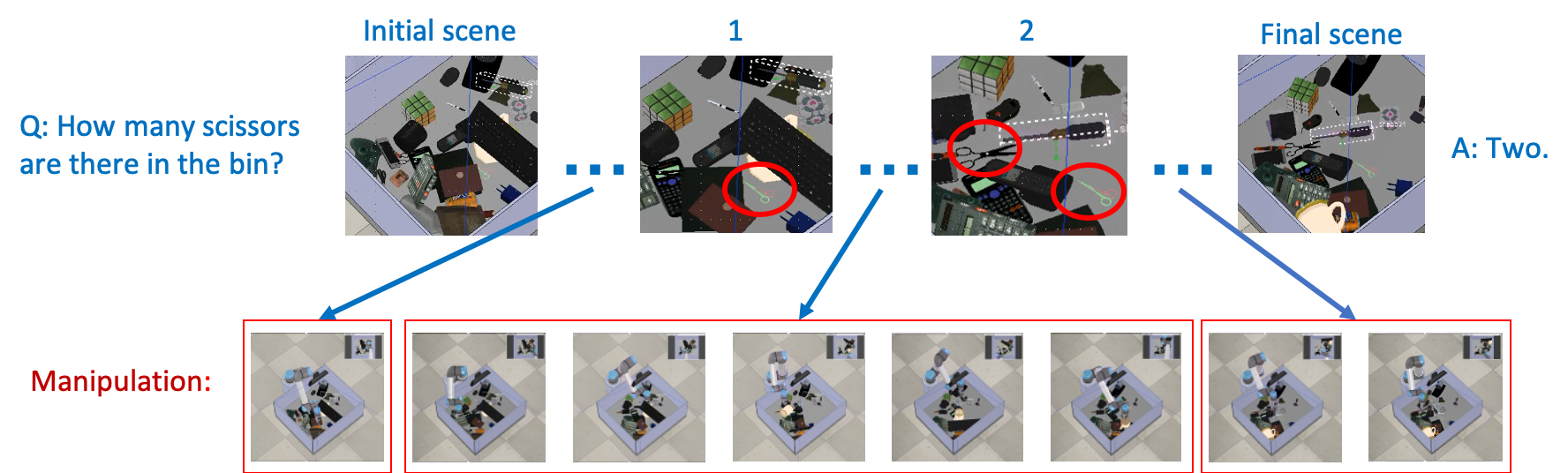}
	\caption{The performance of our MQA system. When the robot receives a question, it will perform a series of manipulation actions to simplify the scene for question answering. After the scene is simple enough for question answering, the robot will output an answer.}
	\label{fig:ex_resu}
\end{figure*}

\section{EXPERIMENTS}
\label{section: experiment}
\subsection{Evaluation of MQA System }
\label{section:evaluation}

In this section, we evaluate the performance of the MQA system. The COUNTING question is specifically selected to solve. The MQA system is evaluated on the test set which includes 30 initial scenes and corresponding question-answer pairs. As the proposed MQA task mainly emphasizes that agent must perform actions to answer a given question, we select a random algorithm, in which both the manipulation action and the number of manipulation are randomly selected as baseline. For the proposed DQN model, we evaluate its performance with three types of reward design. The first system (DQN($R_g$)) is based on our DQN model trained on the reward design of $R_q = R_g$. The second system (DQN($R_l$)) is based on our DQN model trained on the reward design of $R_q = R_l$. The last system (DQN($R_g+R_l$)) is based on our DQN model trained on the reward design of $R_q = 0.5R_g+0.5R_l$.

For the three DQN models, we pretrained them firstly for the manipulation policy learning, and then all the four MQA systems are trained for the answering accuracy improvement (for DQN models, the weights are fixed). Finally, we test the whole system on the COUNTING questions in the dataset. The results are shown in TABLE \ref{table:testing_result}, where we bold the highest accuracy of COUNTING task among the four methods.

\begin{table}[h]
	\centering
	\fontsize{7}{8}\selectfont
	\caption{Accuracy of MQA system with different manipulation policies}
	\begin{tabular}{p{2.5cm}<{\centering}cccc}
		\toprule
		\multicolumn{1}{c}{Method}&
		\multicolumn{1}{c}{Easy Scenes}&
		\multicolumn{1}{c}{Medium Scenes}&
		\multicolumn{1}{c}{Hard Scenes}\cr
		\midrule
		random &0.2771&0.3079&0.2735 \cr
		DQN($R_g$)&\textbf{0.4767}&0.3939&0.3388\cr
		DQN($R_l$)&0.4458&\textbf{0.4242}&\textbf{0.5289}\cr
		DQN($R_g+R_l$)&0.4536&0.4130&0.3838\cr
		\bottomrule
	\end{tabular}
	\label{table:testing_result}
\end{table}

\begin{figure}[b]
	\centering
	\subfigure[the reward obtained in training]{
		\begin{minipage}[t]{0.45\linewidth}
			\centering
			\includegraphics[width=1\linewidth]{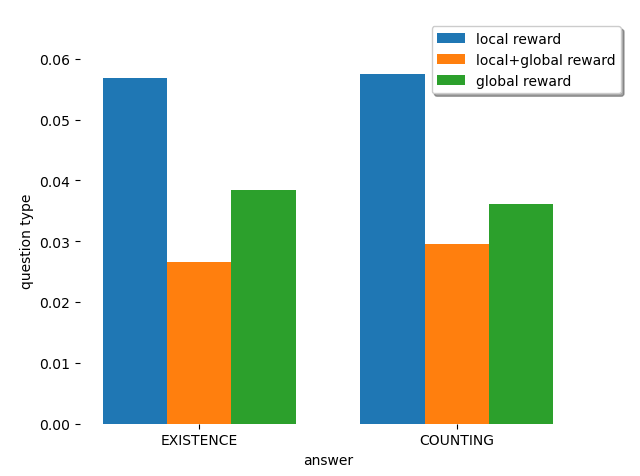}
			\label{fig:training_reward}
		\end{minipage}%
	}%
	\hspace{0.1in}
	\subfigure[the reward obtained in testing]{
		\begin{minipage}[t]{0.45\linewidth}
			\includegraphics[width=1\linewidth]{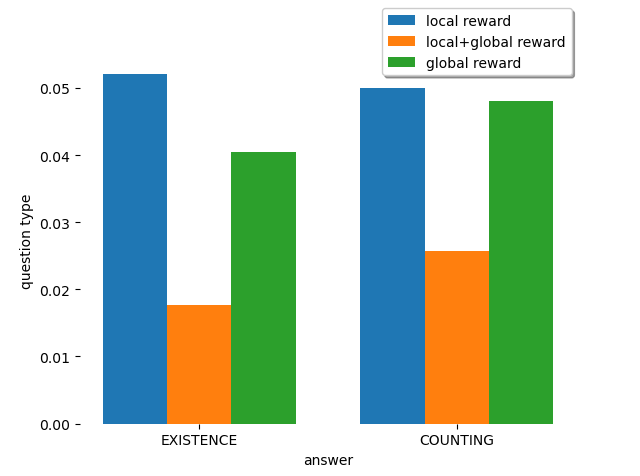}
			\label{fig:testing_reward}
		\end{minipage}%
	}%
	\centering
	\caption{The reward obtained in the training and testing processing }
\end{figure}

As demonstrated in TABLE \ref{table:testing_result}, the accuracy of our MQA system is sometimes less than half. However, compared with the results of counting question in similar tasks such as IQA and EQA, our performance is quite good. Actually, counting question is indeed the most difficult one for this kind of QA tasks. And these results can validate the effectiveness of our framework.

Examples of the experiment are shown in Fig.\ref{fig:ex_resu}. The robot is asked ``How many scissors are there in the bin?'' As is shown, there is no scissor can be seen at the initial scene. And then the robot performs manipulation actions to change the scene. After the robot explores the first area where scissors may be hidden, a pair of scissors is found. But the task is not finished, the exploration continues. After performing several manipulations, the robot finds another pair of scissors. Only when the robot has explored all suspicious areas that may hide scissors, the robot will give the answer. At last, the robot gives the answer ``Two'' after finishing the exploration.

\begin{figure*}[h]
	\centering
	\includegraphics[width=1\linewidth]{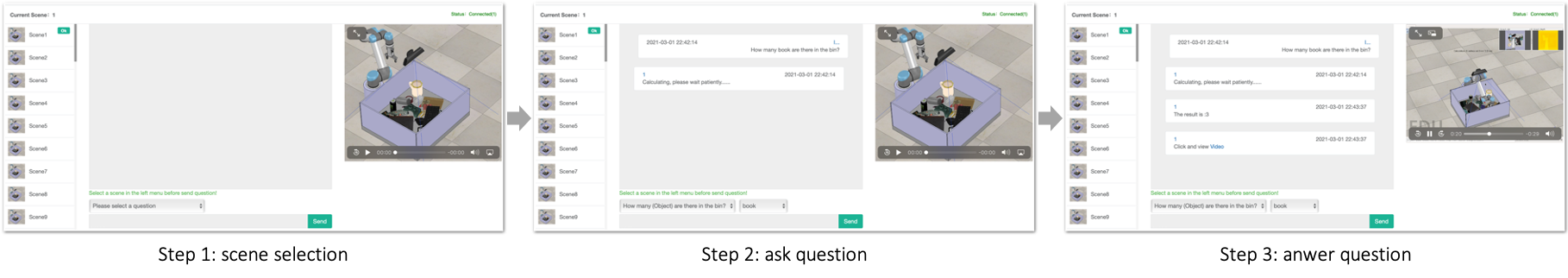}
	\caption{The process of interacting with MQA website system: the user can select scene and ask question. After the robot receives the question, it will explore the scene and then give an answer. The user can receive the video of manipulations and the answer to the question on the website.}
	\label{fig:mqa_demo}
\end{figure*}

\subsection{Ablation Studies on Reward}
\label{ablation_study}
In this section, we further perform the ablation studies on the reward design, the accuracy of MQA system with different manipulation policy is shown in TABLE \ref{table:testing_result}. All of the DQN models perform better than the random baseline in all scenes. DQN($R_l$) system performs best in general, while the DQN($R_g$) system performs worst among the three DQN systems, and the performance of DQN($R_g+R_l$) is between DQN($R_g$) and DQN($R_l$). It is because that the MQA task is a complex task which pays attention to a specific type of target object in the question and needs many steps to accomplish. For a single step, if the robot focuses on the whole scene, the scene will be changed as a whole and the target object will not receive specific attention. If the robot focuses on the target object and takes manipulate to change the scene step by step, the scene will be simplified more efficiently. In addition, we also note that DQN($R_g$) performs best in the easy scene. It is because the number of objects in easy scenes is small enough that the robot can focus on the whole scene to improve efficiency.

Then we analyze the reward obtained in the training process (Fig.\ref{fig:training_reward}) and testing process (Fig.\ref{fig:testing_reward}) based on the three reward designs. It can be seen that the DQN model trained when $R_q = R_l$ or $R_q = R_g$ is likely to get larger reward both in the training and testing process. It is because that the goal of the two models is more specific, either simplifying the whole scene or simplifying the area near the most significant object. The reward obtained when the DQN model is trained on $R_q = R_l$ is the largest both in training and testing process. Therefore, the manipulation policy learned when $R_q = R_l$ is not only the most effective in our task but also the easiest for model learning.


\subsection{Failure Case Analysis}

\begin{figure}[b]
	\centering
	\includegraphics[width=\linewidth]{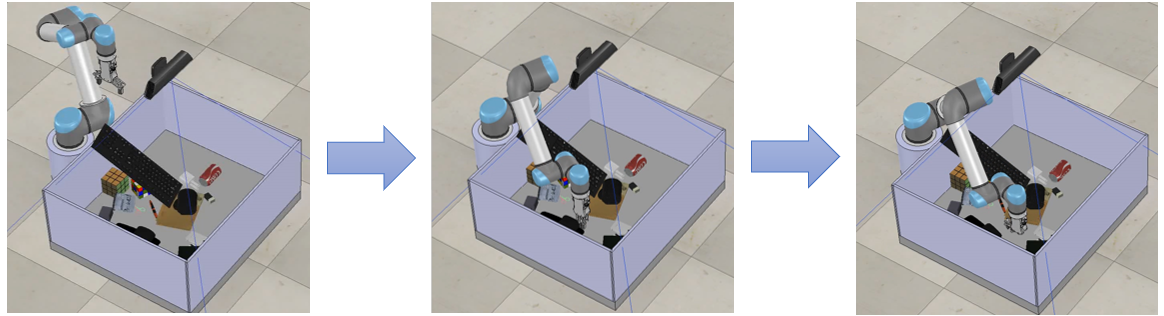}
	\caption{Failure case. When useless manipulation takes place, the state will not be changed and the robot will repeat to generate the useless manipulation.}
	\label{fig:failure}
\end{figure}

Although our model performs well in most situations in MQA task, there are still some failure cases. Comparing to our previous work \cite{deng2019deep}, where the robotic manipulation is implemented to simplify the environment for object grasping, we have solved the problem of generating wrong \textit{stop} manipulation by adding a reward $R_e$ for active exploration in our reward design. But we still cannot solve the problem of generating useless manipulation when the scene is not changed. As Fig.\ref{fig:failure} shows, when useless manipulation takes place, the state will not be changed and the robot will repeat to generate the useless manipulation. We can't guarantee that all manipulations are useful although we have reduced the probability of this situation a lot.

The reason for this situation is that there is no recurrence mechanism in our model, which provides an important alternative for the COUNTING question answering task both for the manipulation module and question answering model. However, when we design our model, we find that if we use recurrence mechanism such as RNN to process a series of image sequences, the efficiency of the model will greatly drop as each picture in the sequence needs to be encoded by CNN. Overall, we would like to propose a lightweight baseline model for this newly proposed MQA task. Therefore, we only take use of the information of the initial image and the final image to improve the efficiency for the question answering task. In our future work, we would like to introduce the recurrence mechanism to the MQA system. For example, we can use LSTM or other algorithm based on RNN to remember the last manipulation taken to output more efficient and non-repetitive manipulations. Another potential improvement is that the question answering model is able to receive more complete information by the recurrence mechanism.

\subsection{MQA Website}


In order to better demonstrate the effectiveness of MQA task, we’ve developed an interactive MQA website which is composed of \textit{Scene Selection} in the left panel, \textit{QA Interface} in the middle panel, and \textit{Video Demonstration} in the right panel. As shown in Fig.\ref{fig:mqa_demo}, it demonstrates the overview of the website and the process of interacting with it. 

To interact with the MQA website, the user can select a scene from the \textit{Scene Selection} module in the first step , which contains 50 different scenes with different objects randomly placed in the bin. After selecting the scene, the scene can be visualized in the \textit{Video demonstration} module. In step 2, the user can ask the question from the option list in the \textit{QA Interface} module. Given the question, the system begins to calculate and generate a sequence of actions for the manipulator. When the calculation is completed, the answer is demonstrated in the \textit{QA Interface} module and it is also possible for the user to watch the process of the manipulation in the \textit{Video Demonstration} module. The interactive MQA website can be reached at \href{http://m-qa.acting-ai.com/}{http://m-qa.acting-ai.com/}.

\section{CONCLUSION}
\label{section: conclusion}
In this paper, we propose a novel task --- Manipulation Question Answering (MQA), where the robot performs manipulation actions to change the environment in order to answer the given question. We have designed a MQA framework that is based on DQN and has specifically solved COUNTING questions. Additionally, a novel dataset that contains a variety of objects, scenes, and corresponding question-answer pairs is established. Extensive experiments have been conducted demonstrating that answering the question via robotic manipulation is highly effective in some practical scenarios. A corresponding benchmark is also established. For the future work, we will further work on other kinds of questions in our dataset and continuously improve the performance of the MQA system. At the same time, we will maintain the interactive MQA website to make it become an open MQA task testing platform. 




\newpage
\begin{thebibliography}{29}
\providecommand{\natexlab}[1]{#1}
\providecommand{\url}[1]{\texttt{#1}}
\expandafter\ifx\csname urlstyle\endcsname\relax
  \providecommand{\doi}[1]{doi: #1}\else
  \providecommand{\doi}{doi: \begingroup \urlstyle{rm}\Url}\fi

\bibitem[Abdul-Kader and Woods(2015)]{abdul2015survey}
Sameera~A Abdul-Kader and JC~Woods.
\newblock Survey on chatbot design techniques in speech conversation systems.
\newblock \emph{International Journal of Advanced Computer Science and
  Applications}, 6\penalty0 (7), 2015.

\bibitem[Anderson et~al.(2018)Anderson, He, Buehler, Teney, Johnson, Gould, and
  Zhang]{anderson2018bottom}
Peter Anderson, Xiaodong He, Chris Buehler, Damien Teney, Mark Johnson, Stephen
  Gould, and Lei Zhang.
\newblock Bottom-up and top-down attention for image captioning and visual
  question answering.
\newblock In \emph{Proceedings of the IEEE Conference on Computer Vision and
  Pattern Recognition}, pages 6077--6086, 2018.

\bibitem[Antol et~al.(2015)Antol, Agrawal, Lu, Mitchell, Batra,
  Lawrence~Zitnick, and Parikh]{antol2015vqa}
Stanislaw Antol, Aishwarya Agrawal, Jiasen Lu, Margaret Mitchell, Dhruv Batra,
  C~Lawrence~Zitnick, and Devi Parikh.
\newblock Vqa: Visual question answering.
\newblock In \emph{Proceedings of the IEEE international conference on computer
  vision}, pages 2425--2433, 2015.

\bibitem[Bajcsy(1988)]{bajcsy1988active}
Ruzena Bajcsy.
\newblock Active perception.
\newblock \emph{Proceedings of the IEEE}, 76\penalty0 (8):\penalty0 966--1005,
  1988.

\bibitem[Batra et~al.(2020)Batra, Chang, Chernova, Davison, Deng, Koltun,
  Levine, Malik, Mordatch, Mottaghi, et~al.]{batra2020rearrangement}
Dhruv Batra, Angel~X Chang, Sonia Chernova, Andrew~J Davison, Jia Deng, Vladlen
  Koltun, Sergey Levine, Jitendra Malik, Igor Mordatch, Roozbeh Mottaghi,
  et~al.
\newblock Rearrangement: A challenge for embodied ai.
\newblock \emph{arXiv preprint arXiv:2011.01975}, 2020.

\bibitem[Bohg et~al.(2013)Bohg, Morales, Asfour, and Kragic]{6672028}
Jeannette Bohg, Antonio Morales, Tamim Asfour, and Danica Kragic.
\newblock Data-driven grasp synthesis—a survey.
\newblock \emph{IEEE Transactions on Robotics}, 30\penalty0 (2):\penalty0
  289--309, 2013.

\bibitem[Bohg et~al.(2017)Bohg, Hausman, Sankaran, Brock, Kragic, Schaal, and
  Sukhatme]{bohg2017interactive}
Jeannette Bohg, Karol Hausman, Bharath Sankaran, Oliver Brock, Danica Kragic,
  Stefan Schaal, and Gaurav~S Sukhatme.
\newblock Interactive perception: Leveraging action in perception and
  perception in action.
\newblock \emph{IEEE Transactions on Robotics}, 33\penalty0 (6):\penalty0
  1273--1291, 2017.

\bibitem[Danielczuk et~al.(2019)Danielczuk, Kurenkov, Balakrishna, Matl, Wang,
  Mart{\'\i}n-Mart{\'\i}n, Garg, Savarese, and
  Goldberg]{danielczuk2019mechanical}
Michael Danielczuk, Andrey Kurenkov, Ashwin Balakrishna, Matthew Matl, David
  Wang, Roberto Mart{\'\i}n-Mart{\'\i}n, Animesh Garg, Silvio Savarese, and Ken
  Goldberg.
\newblock Mechanical search: Multi-step retrieval of a target object occluded
  by clutter.
\newblock In \emph{2019 International Conference on Robotics and Automation
  (ICRA)}, pages 1614--1621. IEEE, 2019.

\bibitem[Das et~al.(2017)Das, Kottur, Gupta, Singh, Yadav, Moura, Parikh, and
  Batra]{das2017visual}
Abhishek Das, Satwik Kottur, Khushi Gupta, Avi Singh, Deshraj Yadav,
  Jos{\'e}~MF Moura, Devi Parikh, and Dhruv Batra.
\newblock Visual dialog.
\newblock In \emph{Proceedings of the IEEE Conference on Computer Vision and
  Pattern Recognition}, pages 326--335, 2017.

\bibitem[Das et~al.(2018{\natexlab{a}})Das, Datta, Gkioxari, Lee, Parikh, and
  Batra]{das2017embodied}
Abhishek Das, Samyak Datta, Georgia Gkioxari, Stefan Lee, Devi Parikh, and
  Dhruv Batra.
\newblock Embodied question answering.
\newblock In \emph{Proceedings of the IEEE Conference on Computer Vision and
  Pattern Recognition Workshops}, pages 2054--2063, 2018{\natexlab{a}}.

\bibitem[Das et~al.(2018{\natexlab{b}})Das, Gkioxari, Lee, Parikh, and
  Batra]{das2018neural}
Abhishek Das, Georgia Gkioxari, Stefan Lee, Devi Parikh, and Dhruv Batra.
\newblock Neural modular control for embodied question answering.
\newblock \emph{arXiv preprint arXiv:1810.11181}, 2018{\natexlab{b}}.

\bibitem[Deng et~al.(2019)Deng, Guo, Wei, Lu, Fang, Guo, Liu, and
  Sun]{deng2019deep}
Yuhong Deng, Xiaofeng Guo, Yixuan Wei, Kai Lu, Bin Fang, Di~Guo, Huaping Liu,
  and Fuchun Sun.
\newblock Deep reinforcement learning for robotic pushing and picking in
  cluttered environment.
\newblock In \emph{2019 IEEE/RSJ International Conference on Intelligent Robots
  and Systems (IROS)}, pages 619--626. IEEE, 2019.

\bibitem[Gandhi et~al.(2020)Gandhi, Gupta, and Pinto]{gandhi2020swoosh}
Dhiraj Gandhi, Abhinav Gupta, and Lerrel Pinto.
\newblock Swoosh! rattle! thump!--actions that sound.
\newblock In \emph{Robotics: Science and Systems}, 2020.

\bibitem[Gordon et~al.(2018)Gordon, Kembhavi, Rastegari, Redmon, Fox, and
  Farhadi]{gordon2017iqa}
Daniel Gordon, Aniruddha Kembhavi, Mohammad Rastegari, Joseph Redmon, Dieter
  Fox, and Ali Farhadi.
\newblock Iqa: Visual question answering in interactive environments.
\newblock In \emph{Proceedings of the IEEE Conference on Computer Vision and
  Pattern Recognition}, pages 4089--4098, 2018.

\bibitem[He et~al.(2017)He, Xia, Yu, Jian, Meng, and Chen]{he2017educational}
Bin He, Meng Xia, Xinguo Yu, Pengpeng Jian, Hao Meng, and Zhanwen Chen.
\newblock An educational robot system of visual question answering for
  preschoolers.
\newblock In \emph{2017 2nd International Conference on Robotics and Automation
  Engineering (ICRAE)}, pages 441--445. IEEE, 2017.

\bibitem[Jang et~al.(2017)Jang, Song, Yu, Kim, and Kim]{jang2017tgif}
Yunseok Jang, Yale Song, Youngjae Yu, Youngjin Kim, and Gunhee Kim.
\newblock Tgif-qa: Toward spatio-temporal reasoning in visual question
  answering.
\newblock In \emph{Proceedings of the IEEE Conference on Computer Vision and
  Pattern Recognition}, pages 2758--2766, 2017.

\bibitem[Johnson et~al.(2017)Johnson, Hariharan, van~der Maaten, Fei-Fei,
  Lawrence~Zitnick, and Girshick]{johnson2017clevr}
Justin Johnson, Bharath Hariharan, Laurens van~der Maaten, Li~Fei-Fei,
  C~Lawrence~Zitnick, and Ross Girshick.
\newblock Clevr: A diagnostic dataset for compositional language and elementary
  visual reasoning.
\newblock In \emph{Proceedings of the IEEE Conference on Computer Vision and
  Pattern Recognition}, pages 2901--2910, 2017.

\bibitem[Kenfack et~al.(2020)Kenfack, Siddiky, Balint-Benczedi, and
  Beetz]{kenfack2020robotvqa}
Franklin~Kenghagho Kenfack, Feroz~Ahmed Siddiky, Ferenc Balint-Benczedi, and
  Michael Beetz.
\newblock Robotvqa—a scene-graph-and deep-learning-based visual question
  answering system for robot manipulation.
\newblock In \emph{IEEE/RSJ International Conference on Intelligent Robots and
  Systems (IROS), Las Vegas, USA}, 2020.

\bibitem[Li et~al.(2016)Li, Hsu, and Lee]{7759839}
Jue~Kun Li, David Hsu, and Wee~Sun Lee.
\newblock Act to see and see to act: Pomdp planning for objects search in
  clutter.
\newblock In \emph{2016 IEEE/RSJ International Conference on Intelligent Robots
  and Systems (IROS)}, pages 5701--5707. IEEE, 2016.

\bibitem[Lu et~al.(2020)Lu, Li, Annamalai, and Yang]{combined2}
Zhenyu Lu, Miao Li, Andy Annamalai, and Chenguang Yang.
\newblock Recent advances in robot-assisted echography: combining perception,
  control and cognition.
\newblock \emph{Cognitive Computation and Systems}, 2\penalty0 (3):\penalty0
  85--92, 2020.

\bibitem[Luo et~al.(2020)Luo, He, and Yang]{combined1}
Jing Luo, Wei He, and Chenguang Yang.
\newblock Combined perception, control, and learning for teleoperation: key
  technologies, applications, and challenges.
\newblock \emph{Cognitive Computation and Systems}, 2\penalty0 (2):\penalty0
  33--43, 2020.

\bibitem[Murahari et~al.(2019)Murahari, Chattopadhyay, Batra, Parikh, and
  Das]{visdial_diversity}
Vishvak Murahari, Prithvijit Chattopadhyay, Dhruv Batra, Devi Parikh, and
  Abhishek Das.
\newblock Improving generative visual dialog by answering diverse questions.
\newblock In \emph{Proceedings of the Conference on Empirical Methods in
  Natural Language Processing (EMNLP)}, 2019.

\bibitem[Nazarczuk and Mikolajczyk(2020)]{nazarczuk2020shop}
Michal Nazarczuk and Krystian Mikolajczyk.
\newblock Shop-vrb: A visual reasoning benchmark for object perception.
\newblock \emph{arXiv preprint arXiv:2004.02673}, 2020.

\bibitem[Rohmer et~al.(2013)Rohmer, Singh, and Freese]{rohmer2013v}
Eric Rohmer, Surya~PN Singh, and Marc Freese.
\newblock V-rep: A versatile and scalable robot simulation framework.
\newblock In \emph{2013 IEEE/RSJ International Conference on Intelligent Robots
  and Systems}, pages 1321--1326. IEEE, 2013.

\bibitem[Sejnova et~al.(2019)Sejnova, Vavrecka, Tesar, and
  Skoviera]{sejnova2019exploring}
Gabriela Sejnova, Michal Vavrecka, Michael Tesar, and Radoslav Skoviera.
\newblock Exploring logical consistency and viewport sensitivity in
  compositional vqa models.
\newblock In \emph{2019 IEEE/RSJ International Conference on Intelligent Robots
  and Systems (IROS)}, pages 2108--2113. IEEE, 2019.

\bibitem[Tan et~al.(2020{\natexlab{a}})Tan, Liu, Guo, Zhang, and
  Sun]{viewpoint}
Sinan Tan, Huaping Liu, Di~Guo, Xinyu Zhang, and Fuchun Sun.
\newblock Towards embodied scene description.
\newblock In \emph{Robotics: Science and Systems}, 2020{\natexlab{a}}.

\bibitem[Tan et~al.(2020{\natexlab{b}})Tan, Xiang, Liu, Guo, and
  Sun]{multiagent}
Sinan Tan, Weilai Xiang, Huaping Liu, Di~Guo, and Fuchun Sun.
\newblock Multi-agent embodied question answering in interactive environments.
\newblock In \emph{Proceedings of the European Conference on Computer Vision
  (ECCV)}, 2020{\natexlab{b}}.

\bibitem[Zeng et~al.(2018{\natexlab{a}})Zeng, Song, Welker, Lee, Rodriguez, and
  Funkhouser]{zeng2018learning}
Andy Zeng, Shuran Song, Stefan Welker, Johnny Lee, Alberto Rodriguez, and
  Thomas Funkhouser.
\newblock Learning synergies between pushing and grasping with self-supervised
  deep reinforcement learning.
\newblock In \emph{2018 IEEE/RSJ International Conference on Intelligent Robots
  and Systems (IROS)}, pages 4238--4245. IEEE, 2018{\natexlab{a}}.

\bibitem[Zeng et~al.(2018{\natexlab{b}})Zeng, Song, Yu, Donlon, Hogan, Bauza,
  Ma, Taylor, Liu, Romo, et~al.]{doi:10.1177/0278364919868017}
Andy Zeng, Shuran Song, Kuan-Ting Yu, Elliott Donlon, Francois~R Hogan, Maria
  Bauza, Daolin Ma, Orion Taylor, Melody Liu, Eudald Romo, et~al.
\newblock Robotic pick-and-place of novel objects in clutter with
  multi-affordance grasping and cross-domain image matching.
\newblock In \emph{2018 IEEE international conference on robotics and
  automation (ICRA)}, pages 3750--3757. IEEE, 2018{\natexlab{b}}.

\end{thebibliography}

\end{document}